\RequirePackage{fix-cm}
\documentclass[twocolumn]{svjour3}          
\smartqed  

\usepackage[utf8]{inputenc} 
\usepackage[T1]{fontenc}    
\usepackage{url}            
\usepackage{booktabs}       
\usepackage{amsfonts}       
\usepackage{nicefrac}       
\usepackage{microtype}      
\usepackage{xcolor}         
\usepackage{soul}
\usepackage{color, xcolor}
\usepackage{epsfig}
\usepackage{graphicx}
\usepackage{amsmath}
\usepackage{amssymb}
\usepackage{multirow}
\usepackage{color, colortbl}
\usepackage{tabularx}
\usepackage[export]{adjustbox}
\usepackage{threeparttable}
\usepackage{pifont}
\usepackage[misc]{ifsym} 
\usepackage{algorithm}
\usepackage{algpseudocode}
\usepackage{paralist}
\usepackage{listings} 
\usepackage{amssymb}
\usepackage{bbding}
\usepackage{url}

\usepackage[
colorlinks=true,
citecolor=blue,         
linkcolor=red,          
CJKbookmarks=True,
pagebackref=true,
]{hyperref}

\usepackage{xspace}
\usepackage{comment}
\usepackage{bm}
\usepackage{wrapfig}

\def\eg{\textit{e.g.}}
\def\ie{\textit{i.e.}}
\def\vs{\textit{vs.\ }}
\def\etc{\textit{etc}}

\newcommand{\green}[1]{{\textcolor[rgb]{0.35,0.79,0.20}{#1}}}
\newcommand{\red}[1]{{\color{red}#1}}

\begin{document}

\sloppy 

\title{Add-SD: Rational Generation without Manual Reference}
\author{
Lingfeng Yang$^{1\dag}$ \and
Xinyu Zhang$^{2,4\dag}$ \and
Xiang Li$^{3\ast}$ \and 
Jinwen Chen$^{4}$ \and
Kun Yao$^{4}$ \and
Gang Zhang$^{4}$ \and
Errui Ding$^{4}$ \and
Lingqiao Liu$^{2}$ \and
Jingdong Wang$^{4}$ \and
Jian Yang$^{1\ast}$
}

\institute{
1. Nanjing University of Science and Technology \\
2. University of Adelaide \\
3. VCIP, CS, Nankai University \\
4. Baidu VIS \\
$\dag$ Equal contributions. $\ast$ Corresponding authors.
}



\maketitle

\begin{abstract}
Diffusion models have exhibited remarkable prowess in visual generalization. Building on this success, we introduce an instruction-based object addition pipeline, named Add-SD, which automatically inserts objects into realistic scenes with rational sizes and positions. Different from layout-conditioned methods, Add-SD is solely conditioned on simple text prompts rather than any other human-costly references like bounding boxes.
Our work contributes in three aspects: proposing a dataset containing numerous instructed image pairs; fine-tuning a diffusion model for rational generation; and generating synthetic data to boost downstream tasks. The first aspect involves creating a RemovalDataset consisting of original-edited image pairs with textual instructions, where an object has been removed from the original image while maintaining strong pixel consistency in the background. These data pairs are then used for fine-tuning the Stable Diffusion (SD) model. Subsequently, the pretrained Add-SD model allows for the insertion of expected objects into an image with good rationale. Additionally, we generate synthetic instances for downstream task datasets at scale, particularly for tail classes, to alleviate the long-tailed problem. Downstream tasks benefit from the enriched dataset with enhanced diversity and rationale.
Experiments on LVIS \texttt{val} demonstrate that Add-SD yields an improvement of 4.3 mAP on rare classes over the baseline. Code and models are available at \url{https://github.com/ylingfeng/Add-SD}.

\end{abstract}

\begin{figure}[t]
    \centering
    \setlength{\fboxrule}{0pt}
    \fbox{\includegraphics[width=0.97\linewidth]{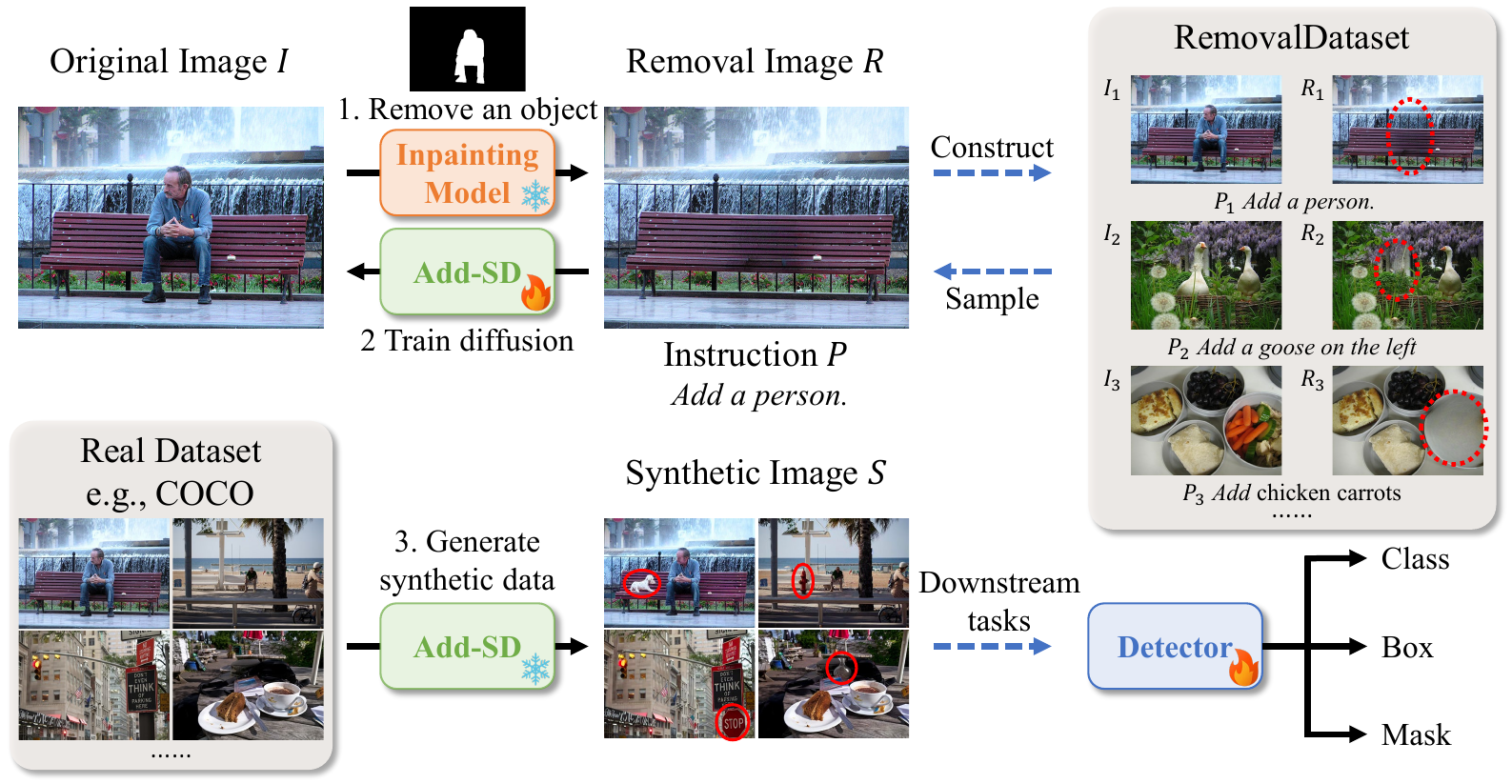}}
    \caption{The proposed Add-SD pipeline begins with the creation of a RemovalDataset containing image pairs via random instance removal. These datasets are then employed to fine-tune image-to-image generation using the Stable Diffusion Model. Next, generation occurs on the entire dataset by sampling rare classes to alleviate the long-tail issue. Finally, synthetic images are integrated into the original dataset to enhance downstream tasks.}
    \label{fig:pipeline_all}
\end{figure}

\section{Introduction}
Generative models~\cite{ho2020denoising,song2020denoising} have shown extraordinary development in vision generalization. Large pretrained diffusion models, \eg, Stable Diffusion (SD)~\cite{rombach2022high}, Imagen~\cite{saharia2022photorealistic}, DALL-E~\cite{ramesh2021zero,ramesh2022hierarchical,betker2023improving} and SDXL~\cite{podell2023sdxl}, provide powerful basis for various specific generation requirements, such as text-to-image editing~\cite{ho2022classifier,rombach2022high}, image-to-image editing~\cite{meng2021sdedit,hertz2022prompt,brooks2023instructpix2pix,ruiz2023dreambooth}, conditional generation~\cite{mou2023t2i,zhang2023adding,li2023photomaker}, and instruction tuning~\cite{dong2023dreamllm,wu2023next,emu2}, \etc. Here, we classify the existing works into two categories, \ie, algorithm-driven methods~\cite{hertz2022prompt,tumanyan2023plug,nichol2021glide,avrahami2022blended,couairon2022diffedit} and data-driven methods~\cite{yang2023paint,li2023gligen,chen2023anydoor,zhao2023x,suri2023gen2det}.

Firstly, among algorithm-driven methods, Prompt-to-Prompt~\cite{hertz2022prompt} and Plug-and-Play~\cite{tumanyan2023plug} utilize activation attention maps between text embeddings and noisy latent feature maps to guide content editing. GLIDE~\cite{nichol2021glide}, Blended Diffusion~\cite{avrahami2022blended}, and DiffEdit~\cite{couairon2022diffedit} manipulate corresponding regions using either manually annotated semantic masks or predictive masks. These approaches excel at editing objects already present in the image. However, when tasked with adding unseen objects, they often struggle to generate desired objects convincingly.

Next, in terms of data-driven methods, Paint-by-Example~\cite{yang2023paint}, GLIGEN~\cite{li2023gligen}, AnyDoor~\cite{chen2023anydoor}, X-Paste~\cite{zhao2023x}, and Gen2Det~\cite{suri2023gen2det} rely on predefined layouts with bounding boxes for image composition. This results in additional manual complexity and restricts layout diversity, thereby limiting generation variety and scalability. Other methods, like InstructPix2Pix~\cite{brooks2023instructpix2pix} and MGIE~\cite{fu2023guiding}, use original-edited image pairs generated via Prompt-to-Prompt~\cite{hertz2022prompt} to edit images with instructions. These approaches are more flexible, with instructions as text prompts being straightforward and avoiding costly layout design. However, it still faces difficulties ensuring consistency of the background content and the rationality of the newly added object.

Based on the literature, there is still no work that addresses the object addition task well, \ie, inserting expected objects into realistic scenes with rational object sizes and positions without manual reference. To achieve this, we propose a simple yet effective pipeline named Add-SD. To ensure practicality and flexibility for human utilization, we follow InstructPix2Pix~\cite{brooks2023instructpix2pix} by simply using instructions to add objects, which effectively reduces the human cost of layout design.

Add-SD consists of three essential stages to complete the object addition task: 1. creating image pairs by removing objects, 2. fine-tuning Add-SD, and 3. generating synthetic data for downstream tasks. {The whole pipeline is depicted in Fig.~\ref{fig:pipeline_all}.}

\begin{figure}[t]
    \centering
    \setlength{\fboxrule}{0pt}
    \fbox{\includegraphics[width=0.97\linewidth]{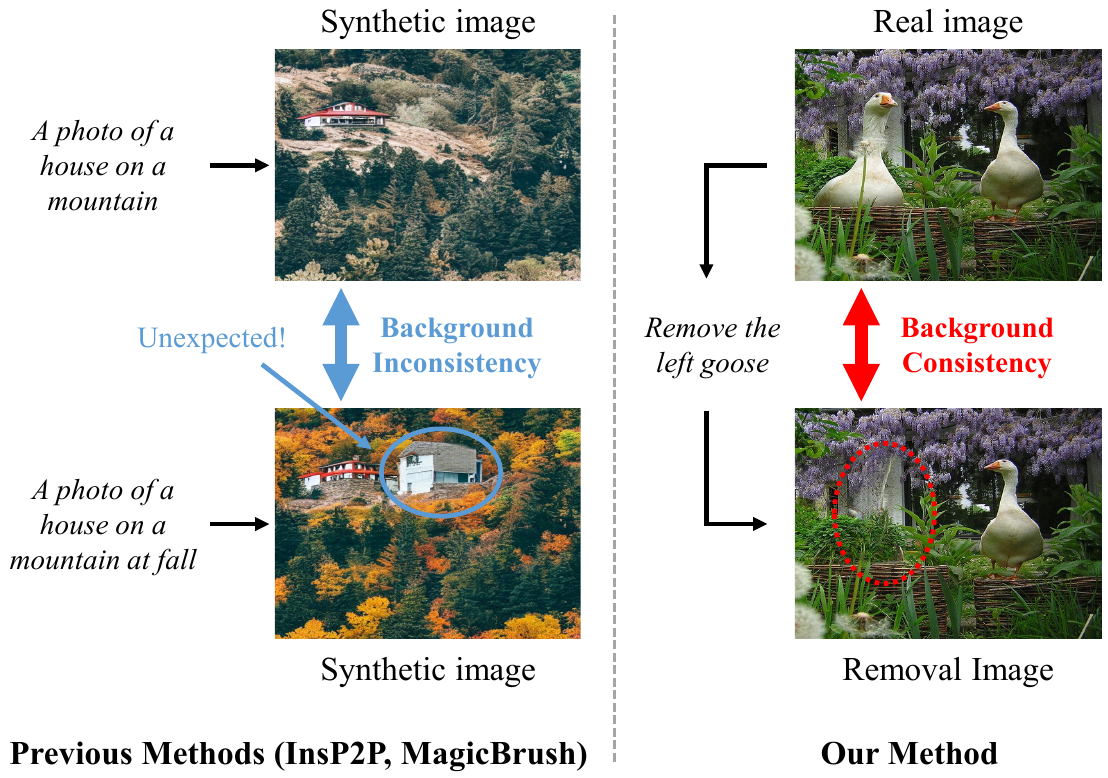}}
    \caption{Our pipeline uses an object removal operation, which ensures that image pairs have a consistent background.}
    \label{fig:house_and_goose}
    \vspace{0pt}
\end{figure}

The first stage aims to construct original-edited image pairs, which serve as the training data for the subsequent instruction-based diffusion model learning on rational object generation. Different previous methods which creates image pairs which are generated from paired-text, we apply reverse thinking by removing specified objects from an real image. This object removal operation ensures that the constructed image pairs have consistent backgrounds, and the removed objects provide the model with examples that satisfy visual plausibility and positional rationality (Fig.~\ref{fig:house_and_goose}). Naturally, since removing and adding objects are inverse operations, the image after removing a specified object can be considered the ``original'' image, while the original image is used as the ``edited'' image in the object generation task. Our objective is to achieve consistent and rational image generations using instructions alone, eliminating the need for manual layout designs. To facilitate this, we develop various instruction templates, such as ``Add a [obj].'' or ``Put the [obj] into the figure,'' which are crafted manually or with support from ChatGPT~\cite{chatgpt}. The placeholder ``[obj]'' represents a detailed description of the expected object, encompassing its category, position, attributes, or relative relationships with other objects. This process results in the creation of RemovalDataset, which consists of original-edited image pairs accompanied by instructions.

The second stage involves training a diffusion model based on the created RemovalDataset, using the removal image and corresponding instruction as image and text conditions, respectively. The de-noising target is the original real image, \ie, after adding the new objects. Using the above-produced original-edited image pairs, we follow InstructPix2Pix~\cite{brooks2023instructpix2pix} to fine-tune a stable diffusion model, transforming it into an instruction-based \emph{addition} stable diffusion model for the object addition task, denoted as the Add-SD model.
After the fine-tuning process, one can use our pretrained model to edit any image by adding objects. By prompting Add-SD with an original image and an instruction denoting the ``[obj]'' with descriptions of the objects expected to be inserted, the model performs the task. The visualization in Fig.~\ref{fig:visualization_comparison} demonstrates that our Add-SD can add the expected objects with rational positions and diverse variations.

In the last stage, we evaluate our Add-SD quantitatively on downstream tasks to show the effectiveness of generated images after adding new objects, as previous synthetic augmentation works~\cite{ghiasi2021simple,zhao2023x,suri2023gen2det} do. Initially, we design a super-label-based sampling strategy to decide which objects should be augmented. Then we employ the grounding detectors to localize the added objects in the generated images. This step also functions as a quality filtering process, wherein low-quality added objects with classification scores below a certain threshold are discarded. After that, we conduct object detection experiments using both real and generated images. Experiments with CenterNet2~\cite{zhou2021probabilistic} on COCO \texttt{val} and LVIS \texttt{val} show that Add-SD brings consistent improvements, especially 4.3 mAP increase on rare classes on the LVIS benchmark. Our Add-SD is a general and flexible method, in which each component can be replaced by state-of-the-art models.

In summary, our contributions are as follows:

\begin{itemize}
\item[$\bullet$] Introduce a flexible method Add-SD, an instruction-based addition pipeline designed to automatically insert objects into real images at rational positions. Add-SD simplifies the process by enabling the use of instructions for object addition without the need for costly human-layout design.

\item[$\bullet$] Propose an instructed image pair creation strategy and form a RemovalDataset by removing objects from an image. This reverse thought process constructs original-edited image pairs, maintaining high visual plausibility and positional rationality.

\item[$\bullet$] Improve the performance of object detection and instance segmentation with a generated synthetic dataset using our proposed Add-SD, where the added objects are localized and filtered by the existing grounding detector.
\end{itemize}

\section{Related Work}

\subsection{Diffusion Models}
Diffusion models for image generation have gained wide attention since the development of large pretrained models, including Stable Diffusion (SD)~\cite{rombach2022high}, Imagen~\cite{saharia2022photorealistic}, DALL-E~\cite{ramesh2021zero,ramesh2022hierarchical,betker2023improving}, and SDXL~\cite{podell2023sdxl}. They are further integrated into specific applications such as personalized image generation~\cite{kawar2023imagic,gal2022image,ruiz2023dreambooth,xiao2023fastcomposer} and local image editing~\cite{hertz2022prompt,brooks2023instructpix2pix,tumanyan2023plug,avrahami2022blended,yu2023inpaint,yang2023paint,chen2023anydoor}.

Personalized image generation methods perform poorly for creating new objects in images, as they generate diverse samples but lack consistency after editing. In local image editing, Prompt-to-prompt~\cite{hertz2022prompt} uses guided attention between prompts and images for training-free editing, while Plug-and-Play~\cite{tumanyan2023plug} enhances attention by focusing on spatial features and self-affinities. However, these methods are limited by their attention guidance, focusing mainly on replacing instances rather than generating completely isolated targets.

Moreover, researchers have discovered that introducing prior masks instead of computed masks in Prompt-to-prompt shows results that better align with the expected generation positions. Blended Diffusion~\cite{avrahami2022blended} employs a multi-step blending approach within the masked region to produce outputs with consistency. Inpaint-Anything~\cite{yu2023inpaint} leverages SAM~\cite{kirillov2023segment} to produce correlated masks. GLIGEN~\cite{li2023gligen} handles grounding control by injecting grounding information into new trainable layers via a gated mechanism. 

Notably, in former works, the added objects are prompted with text instructions, while the image composition task designates the added target from given image conditions. Paint-by-Example~\cite{yang2023paint} inpaints a semantically consistent object derived by CLIP~\cite{radford2021learning} on the figure. 
TF-ICON~\cite{lu2023tf} designs a training-free method to composite cross-domain images.
AnyDoor~\cite{chen2023anydoor} proposes an ID extractor to enhance highly-faithful details. However, the mask-based methods are incompatible with tremendous generation subject to the limitation of requiring manual reference. In contrast, we aim to propose Rational Generation without Manual Reference.

\subsection{Instruction Tuning}
This fine-tuning technique is harnessed to adapt pretrained models using human instructions, with the goal of enhancing usability. It finds extensive application in fine-tuning vision-language models~\cite{liu2024visual,peng2023kosmos,zhang2022glipv2,emu}. Noteworthy examples include LLaVA~\cite{liu2024visual}, which leverages visual instruction tuning through interleaved image-text dialogues with human instructions. Additionally, KOSMOS-2~\cite{peng2023kosmos} integrate bounding boxes into large multimodal models, enabling grounding capabilities.

In the realm of the vision generation~\cite{brooks2023instructpix2pix,hertz2022prompt,zhang2024magicbrush,sheynin2023emu,emu2,koh2024generating}, InstructPix2Pix~\cite{brooks2023instructpix2pix} introduces an image-to-image editing model, harnessing GPT-3~\cite{brown2020language} and Prompt-to-Prompt~\cite{hertz2022prompt} to construct a substantial synthetic dataset for instruction-based image editing. MagicBrush~\cite{zhang2024magicbrush} manually constructs an annotated dataset for instruction-guided image editing. Emu Edit~\cite{sheynin2023emu} employs multi-task training and proposes a matching architecture to address various generation tasks. Moreover, Emu2~\cite{emu2} utilizes a large multimodal model to refine instruction prompting.

While instruction tuning heavily depends on training data pairs and prompted instructions, we highlight the constraints of the synthetic dataset constructed by Prompt-to-Prompt in InstructPix2Pix. Instead, we opt for an inverse operation of removing objects from the scene to generate superior data pairs with high consistency and rational distributions.

\subsection{Synthetic Data as Augmentation}
Several studies have leveraged synthetic data as augmentation to enrich training samples and mitigate overfitting. In general classification tasks~\cite{azizi2023synthetic,he2022synthetic,sariyildiz2023fake}, data generation prompts specific labels from the same domain as the evaluated benchmarks, such as ImageNet~\cite{deng2009imagenet}, and long-tailed datasets like iNaturalist~\cite{van2018inaturalist}. Generating images for classification is simpler than location-based tasks, primarily involving object-centric samples. However, for object detection or instance segmentation, the task complexity increases.

Cut-Paste-and-Learn~\cite{dwibedi2017cut} automatically cuts object instances and pastes them onto random backgrounds. Context-DA~\cite{dvornik2018modeling} improves pasting strategy by modeling surrounding visual context. Then, Copy-Paste~\cite{ghiasi2021simple} found that a simple strategy of randomly picking objects and pasting them at random locations on the target image could provide a significant boost, as long as specialized settings such as large-scale jittering are applied. While the former works use realistic instances cropped from natural images, X-Paste~\cite{zhao2023x} inherits Copy-Paste's strategy but scales it up with more diverse samples generated by diffusion models.

However, these synthetic data struggle with whether to retain the pasted object. The heuristic pasting rule often ignores the rationality of added objects. There is a common understanding that placing training objects at unrealistic positions complicates implicitly modeling context, resulting in reduced detection accuracy~\cite{dwibedi2017cut,dvornik2018modeling}. Instead of complex designs, our Add-SD aims to learn rational addition solely from text instructions with a single diffusion model, offering efficiency and high variability.

\begin{figure*}[t]
    \centering
    \setlength{\fboxrule}{0pt}
    \fbox{\includegraphics[width=0.95\textwidth]{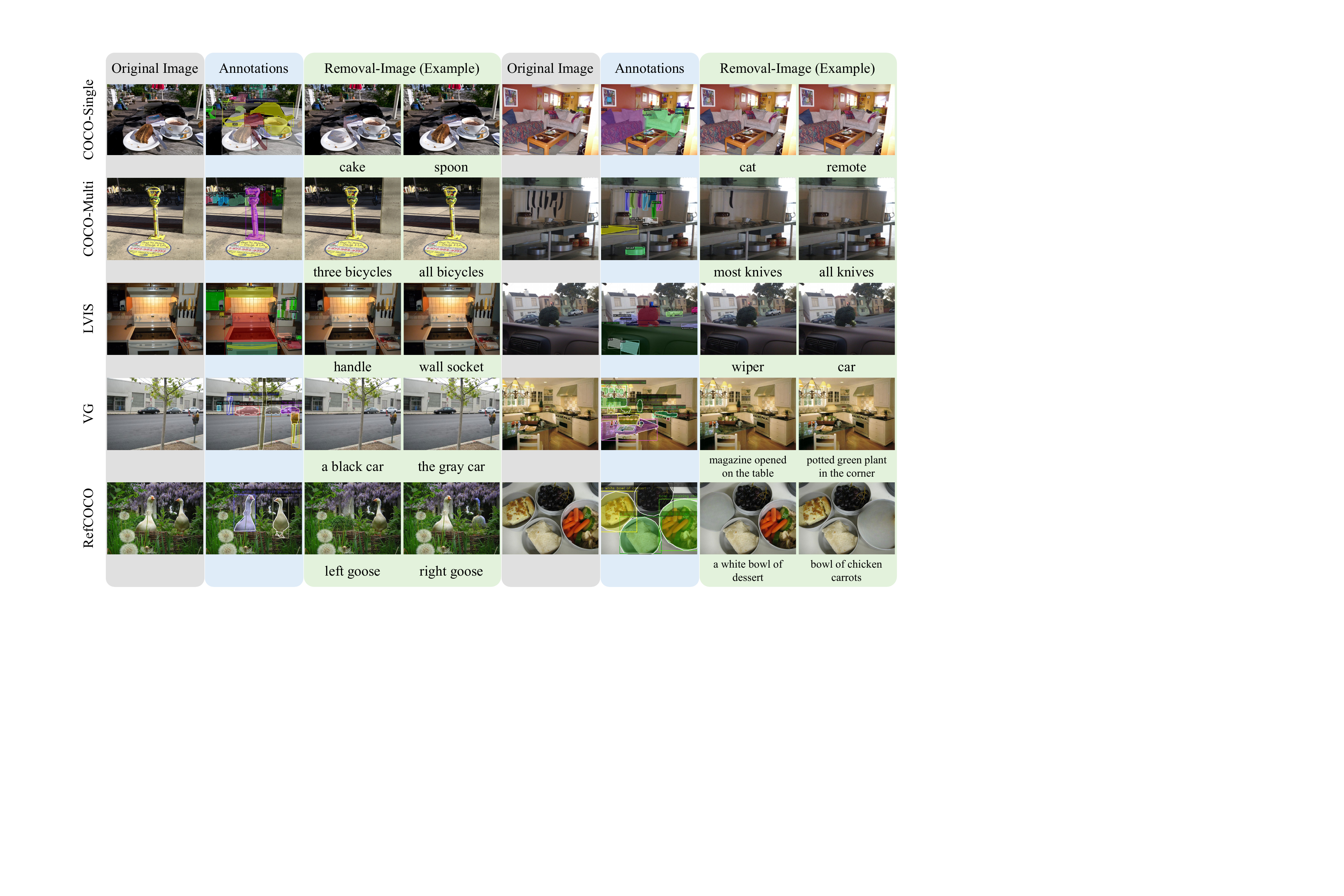}}
    \caption{Visualization of RemovalDataset. The first row shows the removal of a single object from COCO images. The second row involves removing partial instances to facilitate multiple generations. The LVIS dataset is similar to COCO but includes more fine-grained categories. The VG and RefCOCO datasets specify target captions containing attribute and relation information.}
    \label{fig:removal_dataset}
\end{figure*}

\section{Add-SD}

Our work focuses on the object addition task, \ie, inserting isolated objects into real images in a rational manner. This includes believable object placement, natural object size, and seamless blending into the image background without altering the overall layout or obscuring existing foregrounds.

Existing image editing methods address the object addition task through structural designs like attention maps~\cite{hertz2022prompt,ruiz2023dreambooth,tumanyan2023plug,kawar2023imagic}, human reference layouts such as bounding boxes~\cite{nichol2021glide,yang2023paint,li2023gligen,chen2023anydoor}, and instruction-based tuning with data pairs~\cite{brooks2023instructpix2pix,zhang2024magicbrush,sheynin2023emu}. However, they encounter challenges in adding isolated objects, including the inability to independently add targets without reference objects, reliance on manually designed restrictions, and the lack of realistic image pairs that maintain positional and contextual consistency.

We thus propose Add-SD, an instruction-based pipeline specifically designed for the object addition task.
Add-SD can automatically add expected objects based solely on text prompts in a human instruction format, significantly reducing the need for reference layouts such as bounding boxes.
To display a rational generation process for objects' appearances, sizes, and positions, we perform Add-SD in three stages (Fig.~\ref{fig:pipeline_all}): 1. creating image pairs by removing objects, 2. fine-tuning Add-SD, and 3. generating synthetic data for downstream tasks.

\subsection{Creating Image Pairs by Removing Objects}
Image editing involves generating image pairs with consistent content but differing in local regions. Existing methods~\cite{brooks2023instructpix2pix,tumanyan2023plug} first create text pairs with captions for before and after editing, then perform text-to-image generation to obtain synthetic image pairs. Although this approach is straightforward for creating instructed image pairs, ensuring consistency between the two images is challenging because both are generated by models.
To address this, we apply an inverse thinking process by removing existing objects from real images to create image pairs. In the object addition task, the images with removed objects are treated as the ``original'' images, while the real images are considered the ``edited'' images. This approach not only ensures the authenticity of the dataset, which aids in rational learning, but also provides fine-grained control over specifying the areas to be edited.

Here, we leverage LaMa~\cite{suvorov2021resolution}, an inpainting model, with parameters frozen for the object removal process.
Specifically, for each image $I$, we first obtain its instance mask $M = \{ M_1, M_2, ..., M_k \}$ from dataset, where $k$ is the instance number.
If ground-truth masks are available, we use them directly; otherwise, we use the Segment Anything Model (SAM)~\cite{kirillov2023segment} to extract masks from bounding boxes.
Then we use LaMa to remove each object, resulting in $k$ images with removed objects, \ie, $R = \textbf{LaMa}(I, M)$, where $R = \{ R_1, R_2, ..., R_k \}$.
Notably, one can remove multiple instances at a time to enhance the richness of the image pairs and facilitate the diffusion model's ability to generalize across multiple objects.

By performing the removal process, we construct an instructed image-pair dataset, termed RemovalDataset, with each $R_i$ and $I$ as original-edited image pairs and assign corresponding instructions $P$, such as ``Add a [obj].'' 
``[obj]'' typically includes the instance category, as well as descriptions of position, attributes, or relative relationships with other objects.
To enhance user experience and usability, we extend the instruction format using ChatGPT~\cite{chatgpt} to generate various templates, such as ``Put the [obj] into the figure.'' This includes more complex sentences like ``Integrate a [obj] into the scene for enhancement,'' as well as interrogative forms like ``Would you consider inserting a [obj] into the composition?'' For scenarios involving the removal of multiple instances, corresponding instructions might be ``Add two [obj].'' or ``Add [obj$_1$] and [obj$_2$].''

Now, we have constructed numerous original-edited image pairs (Fig.~\ref{fig:removal_dataset}) using the object removal process, preparing them for the subsequent diffusion model fine-tuning. Notably, this image-pair creation pipeline theoretically allows for the generation of infinite samples without restrictions when incorporating existing tools. For instance, any region can be removed from the image, and detectors or large multimodal models can be used to extract object classes or scene descriptions as part of the instruction. For ease of quantitative assessment, we initially create samples using a detection dataset with instances annotated with ground truth descriptions and masks as a pioneering experiment.

\begin{figure*}[t]
    \centering
    \setlength{\fboxrule}{0pt}
    \fbox{\includegraphics[width=0.9\textwidth]{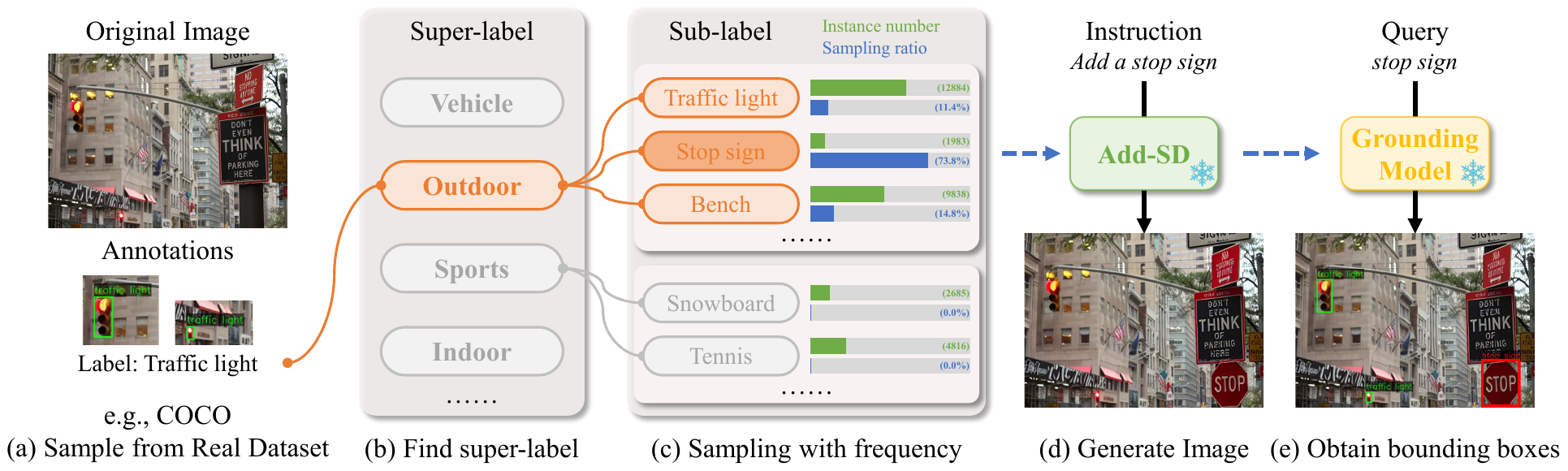}}
    \caption{We design a super-label-based sampling strategy to restrict the category of the added object, ensuring rationality. Then, we randomly sample a sub-label within the super-label, assigning a higher weight to tail-class labels to alleviate the long-tail problem. After image generation, the annotations are inherited from the vanilla dataset \green{(green)} for the original instance and grounded \red{(red)} for the added instance.}
    \label{fig:sample_strategy}
    \vspace{0pt}
\end{figure*}

\subsection{Fine-tuning Add-SD}
In this stage, we fine-tune a diffusion model to learn rational object addition from the above-constructed RemovalDataset. We choose to use the recently popular stable diffusion model~\cite{rombach2022high} as the generative model.
Our model is built upon InstructPix2Pix~\cite{brooks2023instructpix2pix}, \ie, fine-tuning a pretrained SD model conditioned on given image and text instructions.
For an input image $I$, we denote its latent embedding from the VAE encoder $\mathcal{E}$ as $z=\mathcal{E}(I)$.
The image condition is the latent embedding of the removal image $R$ after inputting into $\mathcal{E}$, \ie, $c_I=\mathcal{E}(R)$.
{The text condition is derived by embedding the instructions: $c_T=\mathbf{TextEnc}(P)$.}
Then noise $\epsilon$ sampled from $\mathcal{N}(0, 1)$ is added to $z$ for $t$ times to obtain the noise latent $z_t$:
\begin{equation}
    x_t = \sqrt{\bar{\alpha}_t} x + \sqrt{{1 - \bar{\alpha}_t}} \epsilon,
\end{equation}
where $\bar{\alpha}_t$ is calculated from the accumulated variance schedule and $t$ is sampled from the overall time-steps $T$. Next, a U-Net~\cite{ronneberger2015u} with parameters $\epsilon_\theta$ is utilized to predict the added noise.
$\epsilon_\theta$ is updated via the following optimization objective:
\begin{equation}
    \mathcal{L} = {\left \| {\epsilon - \epsilon_\theta(z_t, t, c_I, c_T)} \right \| }^2.
\end{equation}
Following former practice~\cite{wang2022pretraining,brooks2023instructpix2pix}, we initialize the weights of our model with a pretrained Stable Diffusion checkpoint to maintain its vast text-to-image generation ability.
We do not start our training from InstructPix2Pix pretrained checkpoints, since it is trained on limited synthetic data, which may not align well with images in real scenes and may be insufficient for the object addition task.

After training, the updated model can generate completely new objects in a real image given an adding instruction. For example, if one wants to add a dog to the image, just input ``Add a dog in the image.'' This process is very simple and easy to use. 
The RemovalDataset creation stage and the fine-tuning stage complete the main model of visual generation, \ie, Add-SD. In the next stage, we will use Add-SD to generate a new dataset for downstream tasks to test how these generated images influence performance as a quantitative evaluation.

\subsection{Generating Synthetic Data for Downstream Tasks}
\begin{figure}[t]
    \centering
    \setlength{\fboxrule}{0pt}
    \fbox{\includegraphics[width=0.97\linewidth]{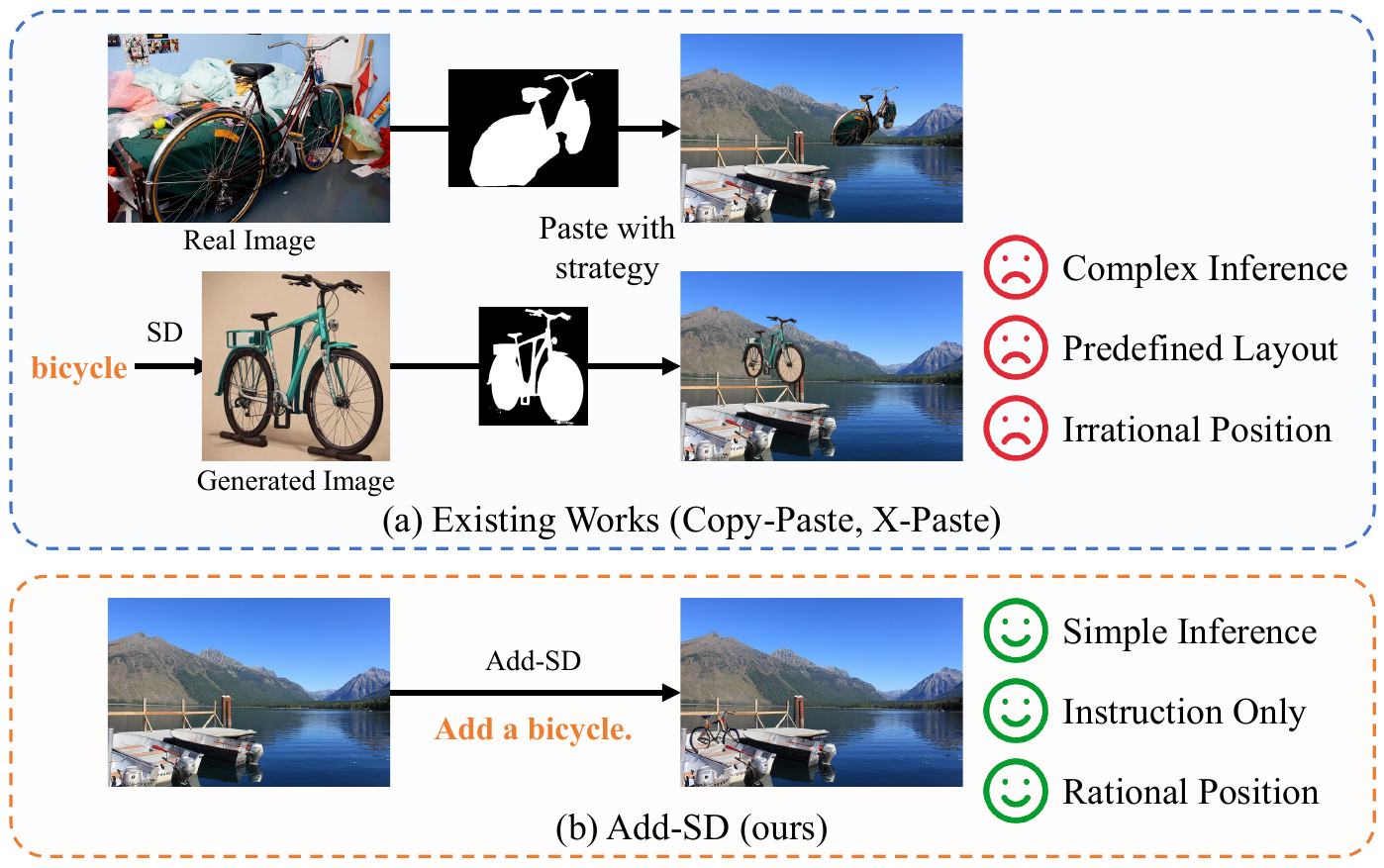}}
    \caption{Copy-Paste~\cite{ghiasi2021simple} and X-Paste~\cite{zhao2023x} pass through a complex synthetic data generation pipeline and may present irrational augmentations. Our Add-SD synthetic augmentation is simple and effective.}
    \label{fig:compare_with_xpaste}
    \vspace{0pt}
\end{figure}

Synthetic data augmentation has been verified as a valuable technique to improve performance in various vision tasks. Since our Add-SD is able to add expected objects, we evaluate its capability on object detection and instance segmentation downstream tasks, especially addressing the long-tail problem. 
Following existing works~\cite{ghiasi2021simple,zhao2023x,suri2023gen2det}, we generate synthetic data based on COCO~\cite{lin2014microsoft} and LVIS~\cite{gupta2019lvis}. Importantly, we consider the following three aspects to generate our new synthetic data and combine it with vanilla data for joint training on detection (Fig.~\ref{fig:sample_strategy}).
Finally, we provide a discussion on the structural comparison between Add-SD and existing works of synthetic data augmentation (Fig.~\ref{fig:compare_with_xpaste}).

\noindent\textbf{Which objects should be added?}
Our Add-SD can flexibly insert specified objects using only text instructions. Therefore, we focus on adding more objects that appear less frequently in the dataset to mitigate the negative effects of the long-tail problem. However, certain categories may be inappropriate for a given image. For instance, it would be implausible to add a dog to an image of a plane in the sky or to include a snowboard in a scene with a traffic light. To address this issue, we propose a method of collecting the super-categories of existing instances within an image, as depicted in Fig.~\ref{fig:sample_strategy} (b). We then sample labels that fall under the same super-category. This approach ensures that the selected category is generated in a rational manner, in principle.
We then sample from the filtered classes, as shown in Fig.~\ref{fig:sample_strategy} (c). Specifically, for COCO, we randomly sample from all classes, with each weight being the reciprocal of the number of label instances. For LVIS, we uniformly sample from the categories officially defined as rare classes. For each image, we add one or multiple rare instances to expand the training sample set.

\begin{figure*}[t]
    \centering
    \setlength{\fboxrule}{0pt}
    \fbox{\includegraphics[width=0.98\textwidth]{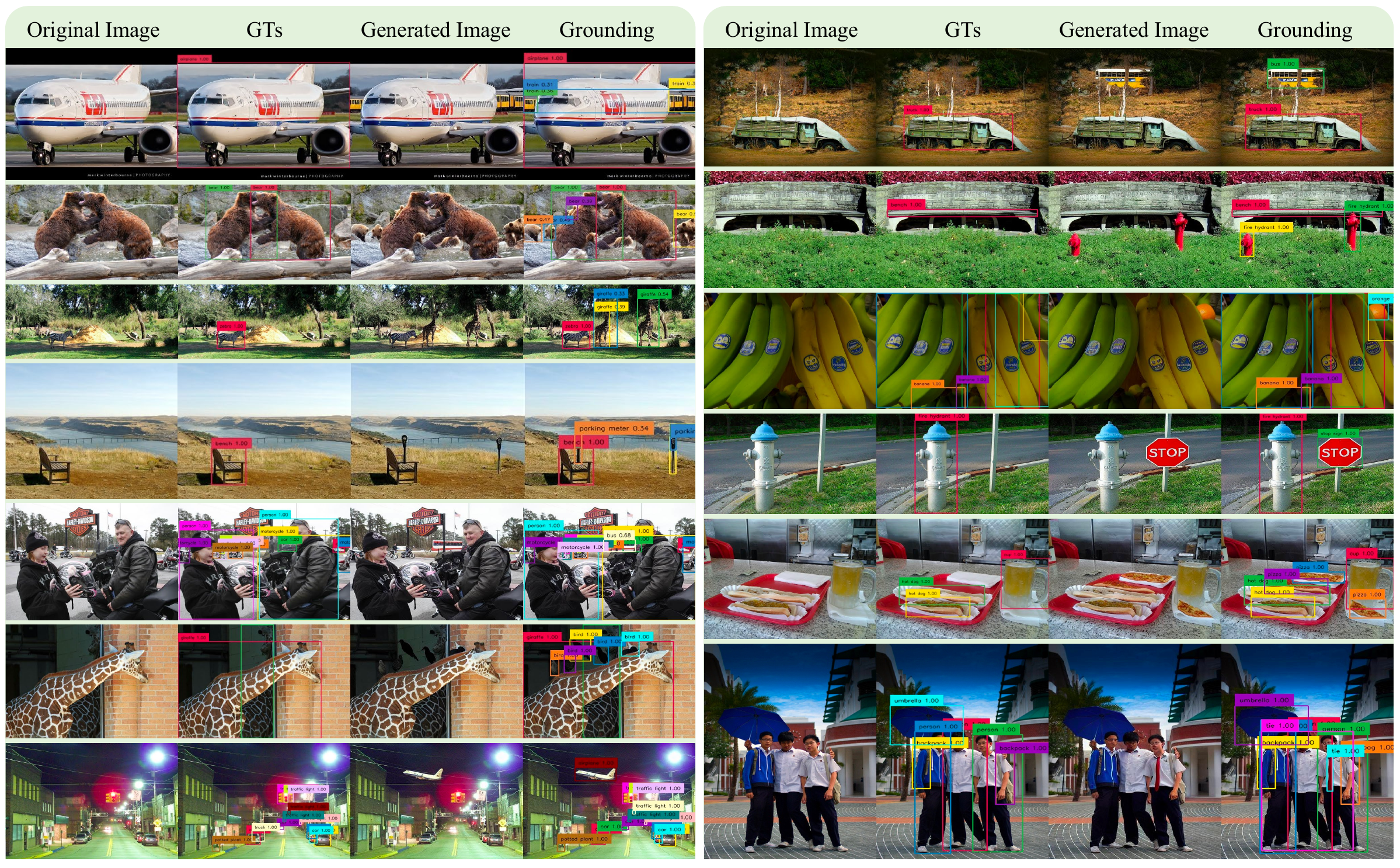}}
    \caption{Generation visualization and grounding results. Add-SD generates diverse synthetic images with new instances, which are then grounded to boxes for training on downstream tasks.}
    \label{fig:generation_and_grounding}
    \vspace{0pt}
\end{figure*}

\noindent\textbf{What types of synthetic images should be used?}
After inserting the expected objects into real images, we obtain various synthetic images. These synthetic images cannot be used for training detectors directly, as we lack annotations for the newly added objects. To address this, we use a grounding method~\cite{liu2023grounding} to locate the added instances, using their labels as queries (see Fig.~\ref{fig:sample_strategy} (e)). 
Notably, the queries include only the class names of the added objects, not all instances in the image. For the original instances, which are expected to be preserved in the synthetic images, we directly copy all ground-truth boxes from the real images into the generated ones.
Additionally, the detection score from grounding results could serves as a data filter, discarding added objects with scores below a specified threshold. We also tested several data filtering strategies from previous works, such as using aesthetic scores or CLIP scores to filter out poor-quality images~\cite{zhao2023x,suri2023gen2det}, using classification scores to remove poor-quality instances originally present in the real images~\cite{suri2023gen2det}, and deleting objects that are either too small or too large~\cite{zhao2023x}. However, these methods did not yield further improvements. Therefore, we have decided to not use any additional filtering strategies for simplicity.
Examples of the generated dataset are depicted in Fig.~\ref{fig:generation_and_grounding}.

\noindent\textbf{How to train the generated data?}
Directly mixing the generated data with the original data using a weighted sampler, \ie, mixing generated images and original images with a specific ratio in a batch, brings limited enhancement. This is because we do not predict masks for the synthetic data; the mask loss is only computed on the real data. This situation causes instability in the gradient updating of mask losses, as it requires multiplying a binary mask to disable calculations on generated samples without a ground truth mask. As a result, we follow Gen2Det~\cite{suri2023gen2det} and use a batch sampling strategy instead, \ie, sampling the entire batch with either original data or synthetic data with a pre-defined threshold. This approach avoids additional processing on the mask loss when the current batch is sampled from synthetic data.

\noindent\textbf{Structural comparison with existing works.}
The synthetic data pipeline of ours and other methods is depicted in Fig.~\ref{fig:compare_with_xpaste}. Copy-Paste~\cite{ghiasi2021simple} and X-Paste~\cite{zhao2023x} both require a segmentor to filter out masks of the target to be pasted. Their difference lies only in whether the target is cropped from a real image or generated by Stable Diffusion~\cite{rombach2022high}. Then, they are pasted onto another real image with strategies such as random pasting or choosing the content with more empty space. Their pipeline is complex and thus time-consuming during the training of detectors. Also, as depicted in Fig.~\ref{fig:compare_with_xpaste} (a), the bicycle may be placed over the lake or in the sky. On the contrary, our method, once trained on an Add-SD with RemovalDataset, requires only one step to generate instances onto it with randomly sampled labels as instructions. Therefore, in Fig.~\ref{fig:compare_with_xpaste} (b), the bicycle is generated on the pier near the boat with a reasonable size and orientation.

\begin{table}[t]
    \centering
    \caption{Human evaluation.}
    \label{tab:methods_comparison_human}
    \resizebox{\linewidth}{!}{
    \begin{tabular}{lccc}
        \toprule
        Method & Quality & Rationality  & Consistency  \\
        \midrule
        InstructPix2Pix & 11.98 & 11.57 & \ \ 6.25 \\ 
        MagicBrush      & 39.18 & 41.42 & 30.11  \\ 
        Add-SD (ours)   & \textbf{48.84} & \textbf{47.01} & \textbf{63.64} \\ 
        \bottomrule
    \end{tabular}}
\end{table}

\section{Experiments}
In this section, we elaborate on the evaluation of Add-SD in terms of image-editing metrics and downstream tasks. First, we summarize the datasets and implementation details of the employed architecture. Then, we evaluate image editing performance through a user study and several generation metrics, comparing it to other commonly used image editing methods. Next, we compare the downstream performance of models trained with our generated data against other synthetic augmentation methods. Additionally, we conduct ablation studies on the components of Add-SD. Finally, we provide qualitative analysis through visualization comparisons.

\subsection{Datasets}
To develop a diffusion model with strong generalization performance, we collect datasets from various fields, including Detection, Visual Question Answering, and Referring Expression Comprehension. The datasets we use are COCO~\cite{lin2014microsoft}, LVIS~\cite{gupta2019lvis}, Visual Genome~\cite{krishna2017visual}, RefCOCO~\cite{yu2016modeling}, RefCOCO+~\cite{yu2016modeling}, and RefCOCOg~\cite{mao2016generation}.

COCO is a widely-used dataset for object detection and segmentation tasks, with over 200,000 images across 80 categories. LVIS expands on COCO with 1,000+ categories and intentionally imbalanced annotations, offering a more challenging benchmark. Visual Genome provides more detailed annotations, including object relationships and attributes, across 100,000+ images. RefCOCO is specialized for referring expression comprehension, offering images with textual descriptions.

In the context of downstream tasks, to ensure fairness, we exclusively utilize COCO as the data source for creating RemovalDataset and generating based on COCO, and similarly for LVIS. Conversely, we train the ultimate Add-SD Model by leveraging all datasets mentioned above, as Visual Genome and RefCOCO offer richer instructions in addition.

\begin{table}[t]
    \centering
    \caption{Quantitative evaluation on image editing tasks.}
    \label{tab:methods_comparison_metrics}
    \resizebox{\linewidth}{!}{
    \begin{tabular}{lccccc}
        \toprule
        Method & L1 $\downarrow$ & L2 $\downarrow$ & CLIP-I $\uparrow$ & CLIP-T $\uparrow$ & DINO $\uparrow$ \\
        \midrule
        InstructPix2Pix & 0.302 & 0.311 & 0.863 & 0.226 & 0.795 \\
        MagicBrush & 0.238 & 0.259 & 0.900 & 0.225 & 0.857 \\
        MGIE &  0.317 &  0.411  & 0.873   &  \textbf{0.227}  &  0.811   \\
        Add-SD (ours) & \textbf{0.196} & \textbf{0.183} & \textbf{0.938} & 0.221 & \textbf{0.928} \\
        \bottomrule
    \end{tabular}}
\end{table}

\subsection{Implementation Details}
For training Add-SD, we follow InstructPix2Pix~\cite{brooks2023instructpix2pix} to fine-tune on RemovalDataset for maximum 50,000 iterations. During inference, we set the diffusion steps $N$ to 100 by default. For downstream tasks, we use the settings of X-Paste~\cite{zhao2023x}, running 4$\times$ scripts with a batch size of 8 per GPU and a total of 90,000 iterations by default. We utilize CenterNet2~\cite{zhou2021probabilistic}, Mask R-CNN~\cite{he2017mask}, and Faster R-CNN~\cite{ren2015faster} with ResNet-50~\cite{he2016deep}, pretrained on the ImageNet-22K, as the detectors. 
For the ablation study, we use CenterNet2 on COCO as the default setting.
The codebase is built upon Detectron2~\cite{wu2019detectron2} and X-Paste~\cite{zhao2023x}. All experiments are conducted on 8$\times$ Tesla V100 GPUs.

\begin{figure*}[t]
    \centering
    \setlength{\fboxrule}{0pt}
    \fbox{\includegraphics[width=0.98\textwidth]{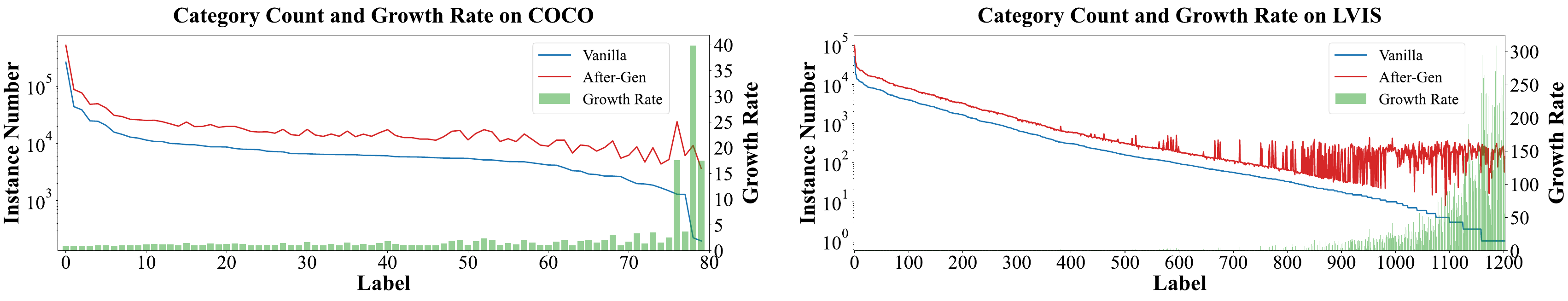}}
    \caption{The instance count for each categories before and after generation and corresponding growth rate over the original on COCO and LVIS datasets.}
    \label{fig:label_count}
\end{figure*}

\begin{table*}[t]
    \caption{Comparisons on LVIS with existing methods using the Centernet2 architecture. The results reveal consistent improvements across various backbones, with particularly higher gains observed on the rare categories.}
    \centering
    \begin{tabular}{lcccccccc}
        \toprule
        \multirow{2.5}{*}{Method}  & \multicolumn{4}{c}{Box} & \multicolumn{4}{c}{Mask} \\
        \cmidrule(lr){2-5} \cmidrule(lr){6-9}
         & AP & AP$_r$ & AP$_c$ & AP$_f$ & AP & AP$_r$ & AP$_c$ & AP$_f$ \\
        \midrule
        Baseline & 33.80 & 20.84 & 32.84 & 40.58 & 29.98 & 18.36 & 29.64 & \textbf{35.46} \\
        Copy-Paste & 34.31 & 21.19 & \textbf{34.32} & 40.07 & 30.29 & 19.97 & 30.43 & 34.67  \\
        X-Paste &34.34 & 21.05 & 33.86 & \textbf{40.71} & 30.18 & 18.77 & 30.11 & 35.28 \\
        Add-SD (ours) & \textbf{34.82} & \textbf{25.11} & 33.90 & 40.11 & \textbf{30.87} & \textbf{22.44} & \textbf{30.44} & 35.05 \\
        \bottomrule
    \end{tabular}
    \label{tab:lvis_comparison}
\end{table*}

\subsection{Image Editing Evaluation}

\noindent\textbf{User Study.}
We generate 100 samples each for InstructPix2Pix~\cite{brooks2023instructpix2pix}, MagicBrush~\cite{zhang2024magicbrush}, and our Add-SD under the same instructions and turn to multiple users ($\sim$10) for blind testing, gathering ratings in terms of visual quality, rationality, and consistency. The quality metric emphasizes the overall visual appeal and fidelity of the generated images. Rationality assesses whether the generated objects make sense or are rational in the visual context, including object size, position, orientation, \etc. Finally, consistency measures how well the background content and overall context are preserved before and after editing. We report the user preference ratio (\%). The results in Table~\ref{tab:methods_comparison_human} show that our method obtains a major preference in human evaluation.

\noindent\textbf{Quantitative metrics for the image editing.}
We employ editing metrics of L1, L2, CLIP-I, CLIP-T, and DINO. L1 and L2 measure pixel-wise distance between generated and ground-truth images from the COCO dataset. CLIP-I and DINO gauge similarity between image features extracted by CLIP and DINO, respectively. CLIP-T evaluates feature similarity between image and text using text input ``Add a [obj]''. Our method outperforms others across most metrics (Table~\ref{tab:methods_comparison_metrics}). Notably, CLIP-T metric has a bias toward only specifying the added object in text, rather than the entire content of the generated image.

\begin{table}[t]
    \caption{Comparisons on COCO with existing methods using Centernet2 architecture.}
    \centering
    \begin{tabular}{lcc}
        \toprule
        Method & Box AP & Mask AP \\
        \midrule
        Baseline &  46.00 & 39.80 \\
        Copy-Paste &  46.40 & 39.80 \\
        X-Paste &  46.60 & 39.90 \\
        Add-SD (ours) & \textbf{46.92} &  \textbf{40.50} \\
        \bottomrule
    \end{tabular}
    \label{tab:coco_comparison}
\end{table}

\subsection{Comparisons on Downstream Tasks}

\noindent\textbf{Different synthetic data augmentation methods.}
We compare the detection and segmentation AP scores between detectors trained with our synthetic data and others using the CenterNet2 architecture for object detection (Box AP) and instance segmentation (Mask AP). We first conduct experiments on the LVIS~\cite{gupta2019lvis} dataset. Note that the official results reported in X-Paste are irreproducible, as stated in the open issue~\footnote{\url{https://github.com/yoctta/XPaste/issues/2}} and the appendix of Gen2Det.
We re-run the LVIS experiments following the exact configuration file of the X-Paste codebase, which aligns with the reproduction results within Gen2Det.
From Table~\ref{tab:lvis_comparison}, we observe that, specifically for the rare classes, our method surpasses the baseline by 4 box AP and X-Paste by 3.8 box AP. Overall, our method demonstrates improvements in box AP/mask AP by +1.0/+0.8 compared to the baseline.
Similarly, Table~\ref{tab:coco_comparison} shows the comparison of various methods on COCO dataset.
Among the methods listed, Add-SD stands out as the most effective, achieving the highest scores for both Box AP and Mask AP. This suggests that the Add-SD method offers significant improvements over existing techniques in terms of accuracy for both object detection and instance segmentation tasks, benefiting from the augmented synthetic datasets. 
Additionally, as shown in Fig.~\ref{fig:label_count}, the rare classes generated through Add-SD help alleviate the long-tail problem, as the increased number of rare classes leads to a higher growth rate.
Consequently, Add-SD presents a promising approach for enhancing performance in these domains when utilizing the CenterNet2 architecture on the LVIS and COCO datasets.

\begin{table}[t]
    \caption{Comparisons on COCO with different backbones.}
    \centering
    \begin{tabular}{lcccc}
        \toprule
        Method & AP &  AP$_s$ & AP$_m$ & AP$_l$ \\
        \midrule
        Faster R-CNN & 40.20 &  23.96 & 43.47 & \textbf{51.87} \\
        Add-SD & \textbf{40.35}  & \textbf{24.40} & \textbf{43.95} & 51.85 \\
        \hline
        Mask R-CNN & 41.25  & 25.25 & 44.35 & \textbf{53.87} \\
        Add-SD &  \textbf{41.30} & \textbf{25.27} & \textbf{44.60} & 53.31 \\
        \hline
        Centernet2 &  46.00 & -- & -- & -- \\
        Add-SD & \textbf{46.92}  & 31.04 & 51.09 & 59.26 \\
        \bottomrule
    \end{tabular}
    \label{tab:backbone_comparison_coco}
\end{table}

\begin{table*}[t]
    \caption{Ablation on the sampling ratio on COCO.}
    \centering
    \begin{tabular}{ccccccccccccc}
        \toprule
        \multirow{2.5}{*}{Ratio (\%)} & \multicolumn{6}{c}{Box} & \multicolumn{6}{c}{Mask} \\
        \cmidrule(lr){2-7} \cmidrule(lr){8-13}
         & AP & AP$_{50}$ & AP$_{75}$ & AP$_s$ & AP$_m$ & AP$_l$ & AP & AP$_{50}$ & AP$_{75}$ & AP$_s$ & AP$_m$ & AP$_l$ \\
        \midrule
        0 & 46.00 & -- & -- & -- & -- & -- & 39.80 & -- & -- & -- & -- & -- \\
        10 & 46.71 & 65.73 & 50.57 & 31.24 & 51.09 & 59.04 & 40.48 & \textbf{62.56} & 43.71 & 22.87 & 44.44 & 55.61 \\
        20 & \textbf{46.92} & \textbf{65.80} & \textbf{50.77} & 31.04 & 51.09 & \textbf{59.26} & \textbf{40.50} & 62.55 & \textbf{43.73} & 22.97 & \textbf{44.52} & \textbf{55.62} \\
        30 & 46.66 & 65.52 & 50.47 & \textbf{31.66} & 51.07 & 58.44 & 40.26 & 62.24 & 43.48 & \textbf{23.00} & 44.33 & 55.51 \\
        50 & 46.51 & 65.50 & 50.32 & 31.26 & \textbf{51.33} & 58.39 & 40.00 & 62.09 & 43.14 & 22.56 & 44.30 & 54.63  \\
        \bottomrule
    \end{tabular}
    \label{tab:sampling_ratio}
    \vspace{0pt}
\end{table*}

\begin{table*}[t]
    \caption{Ablation on the sampling strategy on COCO.}
    \vspace{0pt}
    \centering
    \resizebox{\textwidth}{!}{
    \begin{tabular}{lcccccccccccc}
        \toprule
        \multirow{2.5}{*}{Strategy} & \multicolumn{6}{c}{Box} & \multicolumn{6}{c}{Mask} \\
        \cmidrule(lr){2-7} \cmidrule(lr){8-13}
         & AP & AP$_{50}$ & AP$_{75}$ & AP$_s$ & AP$_m$ & AP$_l$ & AP & AP$_{50}$ & AP$_{75}$ & AP$_s$ & AP$_m$ & AP$_l$ \\
        \midrule
        Normal Mixture & 46.71 & 65.73 & 50.57 & \textbf{31.24} & \textbf{51.09} & 59.04 & 40.48 & \textbf{62.56} & 43.71 & 22.87 & 44.44 & 55.61 \\
        Batch Sampling & \textbf{46.92} & \textbf{65.80} & \textbf{50.77} & 31.04 & \textbf{51.09} & \textbf{59.26} & \textbf{40.50} & 62.55 & \textbf{43.73} & \textbf{22.97} & \textbf{44.52} & \textbf{55.62} 
        \\
        \bottomrule
    \end{tabular}
    }
    \label{tab:batch_sampling}
    \vspace{0pt}
\end{table*}

\begin{table*}[t]
    \caption{Comparisons on COCO under CenterNet2 on the training schedule varying from 4$\times$ to 8$\times$.}
    \centering
    \resizebox{\textwidth}{!}{
    \begin{tabular}{clcccccccccccc}
        \toprule
        \multirow{2.5}{*}{Schedule} & \multirow{2.5}{*}{Method} & \multicolumn{6}{c}{Box} & \multicolumn{6}{c}{Mask} \\
        \cmidrule(lr){3-8} \cmidrule(lr){9-14}
         &  & AP & AP$_{50}$ & AP$_{75}$ & AP$_s$ & AP$_m$ & AP$_l$ & AP & AP$_{50}$ & AP$_{75}$ & AP$_s$ & AP$_m$ & AP$_l$ \\
        \midrule
        \multirow{2}{*}{4$\times$} & Baseline & 46.00 & -- & -- & -- & -- & -- & 39.80 & -- & -- & -- & -- & -- \\
         & Add-SD (ours) & \textbf{46.92} & 65.80 & 50.77 & 31.04 & 51.09 & 59.26 & \textbf{40.50} & 62.55 & 43.73 & 22.97 & 44.52 & 55.62 \\
        \hline
        \multirow{2}{*}{8$\times$} & Baseline & 46.56 & 65.80 & 50.19 & \textbf{32.82} & 50.48 & 58.50 & 40.36 & 62.64 & 43.32 & \textbf{23.90} & 44.18 & 54.70  \\
         & Add-SD (ours) & \textbf{47.22} & \textbf{66.56} & \textbf{51.13} & 32.43 & \textbf{51.21} & \textbf{59.56} & \textbf{40.84} & \textbf{63.19} & \textbf{44.05} & 23.40 & \textbf{45.04} & \textbf{54.85}  \\
        \bottomrule
    \end{tabular}
    }
    \label{tab:training_schedule}
    \vspace{0pt}
\end{table*}

\noindent\textbf{Different detectors.}
To verify the transferability of generated datasets, we additionally compare our methods against the baseline of Faster R-CNN~\cite{ren2015faster}, Mask R-CNN~\cite{he2017mask}, and CenterNet2~\cite{zhou2021probabilistic} benchmarked with COCO. We use the codebase and configs in Detectron2~\cite{wu2019detectron2} with a 3$\times$ schedule using ResNet-50 and FPN settings. From Table~\ref{tab:backbone_comparison_coco}, we observe a consistent improvement of Add-SD over the baseline, indicating the robustness of our generalized data. Notably, on CenterNet2, it achieves the largest performance boost of 0.85 points.

\subsection{Ablation Study}
\noindent\textbf{Synthetic data sampling ratio.}
Using complete generation data for training may cause severe performance drop due to a wide distribution change from the real images. A common solution is to mix the synthetic data and real data for joint training. Therefore, we specifically explore the mixture ratio between the two datasets. The ratio represents the proportion of generated images to all training samples, where a ratio of 0 equals the baseline and a ratio of 1 means using synthetic data only. 
Results in Table~\ref{tab:sampling_ratio} indicate that a sampling ratio of 20\% yields the best performance for mixed learning on COCO. To maintain consistency, the same ratio is applied to the LVIS dataset.
Notably, a mixture ratio of 50\% still improves the baseline by 0.5 box AP, demonstrating the effectiveness of the generated training data.

\noindent\textbf{Batch sampling.}
In this section, we ablate the sampling strategy. Table~\ref{tab:batch_sampling} compares the performance of two sampling strategies: ``Normal Mixture'' and ``Batch Sampling.'' The ``Normal Mixture'' strategy samples both real and synthetic images within a single batch, while the ``Batch Sampling'' strategy samples the two types of images in separate batches.
Across the evaluation on COCO using CenterNet2, ``Batch Sampling'' demonstrates better results, suggesting that it is a viable enhancement strategy. This approach potentially offers benefits in terms of computational efficiency or ease of optimization.

\noindent\textbf{Training schedule.}
We explore the influence of the training schedule on performance by extending the training time from 4$\times$ to 8$\times$ scripts. Table~\ref{tab:training_schedule} shows that our method achieves continuous improvement with the extended training time. Benefiting from the synthetic data, the detector can prevent overfitting and boost performance in scaled-up settings. Notably, Gen2Det~\cite{suri2023gen2det} generates 5$\times$ more instances than ours. However, by training for 2$\times$ longer, we can achieve better performance than Gen2Det (47.22 \vs 47.18 box AP).

\begin{table*}[t]
    \caption{Results on COCO under CenterNet2 comparing the label selection strategy during data generation of Add-SD.}
    \centering
    \resizebox{\linewidth}{!}{
    \begin{tabular}{lcccccccccccc}
        \toprule
        \multirow{2.5}{*}{Strategy} & \multicolumn{6}{c}{Box} & \multicolumn{6}{c}{Mask} \\
        \cmidrule(lr){2-7} \cmidrule(lr){8-13}
         & AP & AP$_{50}$ & AP$_{75}$ & AP$_s$ & AP$_m$ & AP$_l$ & AP & AP$_{50}$ & AP$_{75}$ & AP$_s$ & AP$_m$ & AP$_l$ \\
        \midrule
        w/o Super-label & 46.65 & 65.48 & 50.48 & 30.97 & 50.93 & 58.93 & 40.17 & 62.09 & 43.40 & 22.81 & 44.05 & 55.10 \\
        w/ Super-label & \textbf{46.92} & \textbf{65.80} & \textbf{50.77} & \textbf{31.04} & \textbf{51.09} & \textbf{59.26} & \textbf{40.50} & \textbf{62.55} & \textbf{43.73} & \textbf{22.97} & \textbf{44.52} & \textbf{55.62} \\
       \bottomrule
    \end{tabular}
    }
    \label{tab:super_label_freq}
\end{table*}

\begin{table*}[t]
    \caption{Comparisons on the scale of synthetic datasets on LVIS.}
    \centering
    \resizebox{0.76\linewidth}{!}{
    \begin{tabular}{ccccccccc}
        \toprule
        \multirow{2.5}{*}{Synthetic Data Scale}  & \multicolumn{4}{c}{Box} & \multicolumn{4}{c}{Mask} \\
        \cmidrule(lr){2-5} \cmidrule(lr){6-9}
         & AP & AP$_r$ & AP$_c$ & AP$_f$ & AP & AP$_r$ & AP$_c$ & AP$_f$ \\
        \midrule
        100K & 34.60 & 24.05 & 33.60 & \textbf{40.36} & 30.69 & 22.12 & 30.06 & \textbf{35.15}  \\
        200K & \textbf{34.82} & \textbf{25.11} & \textbf{33.90} & 40.11 & \textbf{30.87} & \textbf{22.44} & \textbf{30.44} & 35.05 \\
        \bottomrule
    \end{tabular}
    }
    \label{tab:synthetic_data_scale}
\end{table*}

\begin{figure*}[t]
    \centering
    \setlength{\fboxrule}{0pt}
    \fbox{\includegraphics[width=0.95\textwidth]{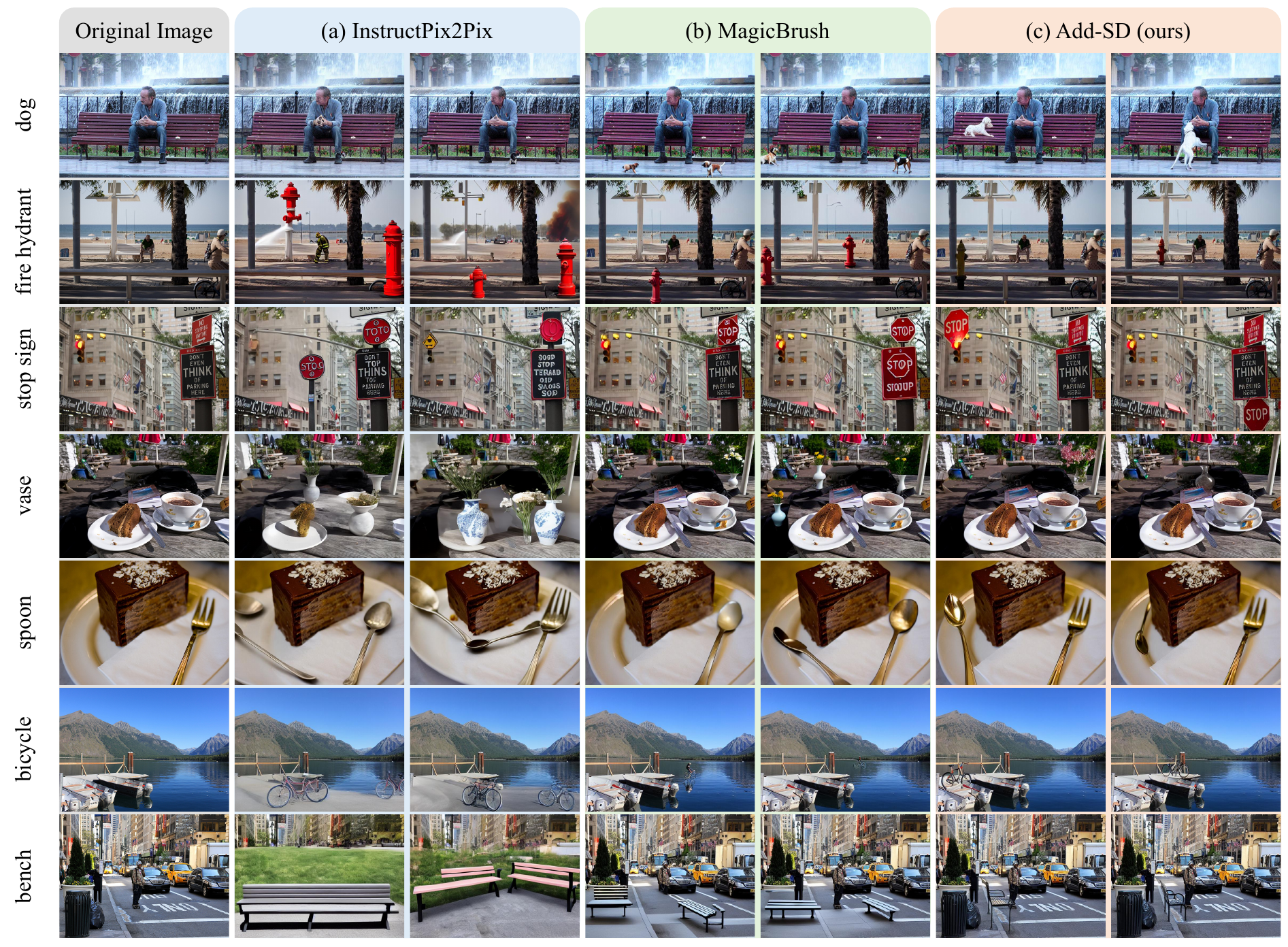}}
    \caption{Comparisons of Add-SD and other image editing methods on generation results with various objects.}
    \label{fig:visualization_comparison}
    \vspace{-3pt}
\end{figure*}

\noindent\textbf{Super-label selection strategy.}
In order to generate samples for rare classes to alleviate the long-tailed problem, we prompt Add-SD with labels under label-wise frequency calculated from COCO. Moreover, we observe that certain labels cannot be generated for specific figures. For example, it is difficult to generate a dog in an image capturing a plane in the sky. Therefore, we propose a label selection strategy by selecting labels in the range of the super-labels of the original instances in the images. In this case, the generated instances and the original instances in the image exhibit relatively similar characteristics, reducing the occurrence of unreasonably generated instances. Table~\ref{tab:super_label_freq} shows that using super-labels demonstrates an effective enhancement in detection performance by generating more reasonable data samples.

\begin{figure*}[t]
    \centering
    \setlength{\fboxrule}{0pt}
    \fbox{\includegraphics[width=0.95\textwidth]{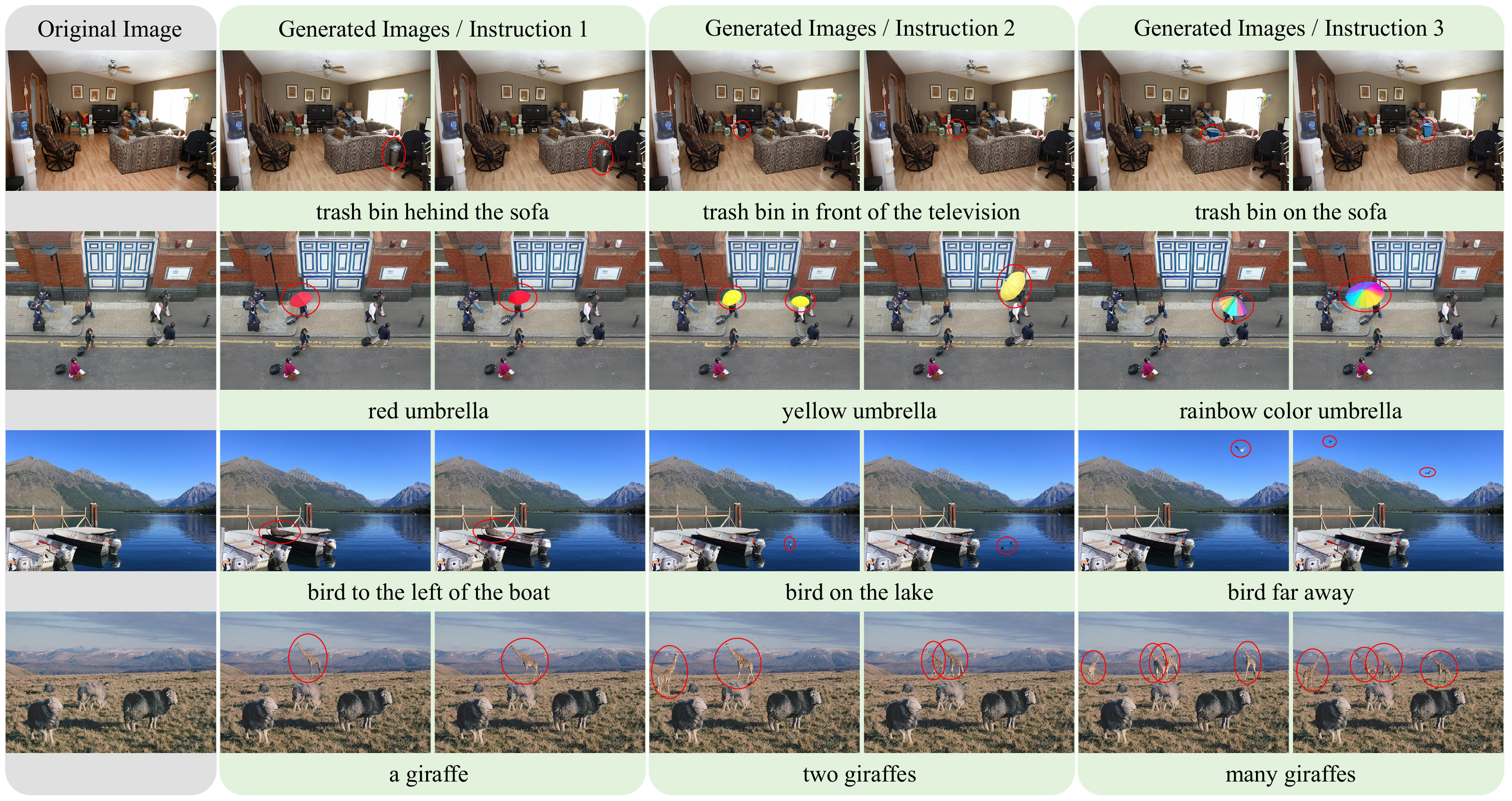}}
    \caption{Visualizations of Add-SD-generated results under rich instructions.}
    \label{fig:more_vis_addsd}
    \vspace{-3mm}
\end{figure*}

\begin{figure}[t]
    \centering
    \setlength{\fboxrule}{0pt}
    \fbox{\includegraphics[width=0.97\linewidth]{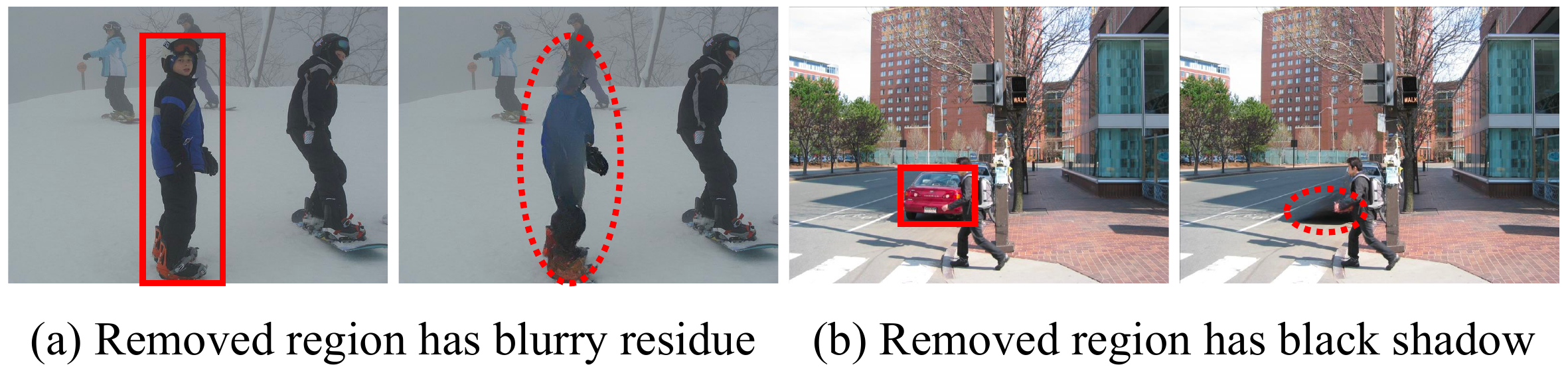}}
    \vspace{0pt}
    \caption{Bad cases of RemovalDataset.}
    \label{fig:bad_case}
    \vspace{-5mm}
\end{figure}

\noindent\textbf{Scale up the synthetic data.}
To indicate the generative diversity of Add-SD, we scale the total number of generated samples from 100K to 200K images. Notably, our total number of iterations remains unchanged (90K) to ensure a fair comparison.
Table~\ref{tab:synthetic_data_scale} illustrates the scalability of synthetic data augmentation with samples generated by Add-SD in downstream tasks, highlighting the benefits derived from diverse training samples.

\subsection{Qualitative Results}
\noindent\textbf{Compare to different image editing methods.}
In this section, we compare the visualization of the editing results between our Add-SD and other image editing methods including InstructPix2Pix~\cite{brooks2023instructpix2pix}, and MagicBrush~\cite{zhang2024magicbrush}.  
From Fig.~\ref{fig:visualization_comparison}, we observe that our method maintains background consistency while providing diverse generalization. For example, in the first image, we generate dogs with different poses and create fire hydrants with various locations and colors, whereas other methods produce similar samples. Additionally, MagicBrush tends to replace existing objects with generated ones, resulting in the spoon covering the original fork. Furthermore, existing methods fail to ensure rationality in the position of the bicycle or the sizes of the dog.

\noindent\textbf{Learn from instruction.}
Since we use datasets like RefCOCO and Visual Genome, which include descriptive statements as instructions, the diffusion model can learn about object properties, orientations, and their relative relationships with surrounding objects. This enables more controlled and rational generation. By managing the instruction inputs, we can produce more diverse images with reasonable placement of new objects.
For each image in Fig.~\ref{fig:more_vis_addsd}, we design different instructions to show the generation diversity of Add-SD. The first line shows the relation awareness of Add-SD, which can distinguish the surrounding objects in complex environments such as the living room. The second sample shows Add-SD can handle attributes as inputs for generating more detailed instances. The example shown in the third line demonstrates a perspective relationship, with near objects appearing larger and distant objects appearing smaller. It also illustrates the ability to discern positional relationships between objects. Finally, we show that Add-SD can generate different numbers of targets by prompting with direct numbers or quantity descriptors. It is notable that when we prompt with ``Add 2 giraffes'', roughly 3/4 of the images (with a total generation of 50 trials) successfully generated two objects. However, in other cases, it's still possible to generate one or more instances, indicating that instructions regarding quantity remain a challenging issue to address.

\noindent\textbf{Limitations.}
There are some limitations to Add-SD. Since the training datasets are generated using an inpainting model, they can be affected by cases where the object is not removed naturally, such as residual blur or shadows (see Fig.~\ref{fig:bad_case}). Additionally, processing large targets is more challenging, as removing a large instance can result in unsatisfactory inpainting, which reduces dataset quality.

\section{Conclusion}
We propose Add-SD, a novel visual generation method for instruction-based object addition, demonstrating significant advancements in seamlessly integrating objects into realistic scenes using only textual instructions. By employing object removal, we create instructed image pairs and construct a dataset called RemovalDataset. We then train an instruction-based stable diffusion (SD) model on this dataset to learn diverse and rational object generation without relying on costly human annotations such as bounding boxes.
The fine-tuned Add-SD enhances image generation quality, rationality, and background consistency, as verified through user studies and various editing metrics. Additionally, we design a super-label sampling strategy to generate synthetic samples to boost downstream tasks. Evaluation on the LVIS \texttt{val} datasets reveals a notable improvement of 4.3/1.0 AP points on rare classes/overall compared to the baseline, highlighting the effectiveness of Add-SD. This approach holds promise for applications in various domains requiring automatic scene augmentation.

\section*{Acknowledgment}
The authors would like to thank the editor and the anonymous reviewers for their critical and constructive comments and suggestions. 
This work is supported by
the National Science Fund of China under Grant No. 62361166670,
the Young Scientists Fund of the National Natural Science Foundation of China (Grant No. 62206134), 
the Fundamental Research Funds for the Central Universities (070-63233084), 
and the Tianjin Key Laboratory of Visual Computing and Intelligent Perception (VCIP). 
Note that Lingfeng Yang and Jian Yang are with 
PCA Lab, Key Lab of Intelligent Perception and Systems for High-Dimensional Information of Ministry of Education, and Jiangsu Key Lab of Image and Video Understanding for Social Security, School of Computer Science and Engineering, Nanjing University of Science and Technology.

\section*{Broader Impacts}
Further research and careful consideration are necessary when utilizing this technology, as the presented proposed method relies on statistics derived from training datasets that may possess biases and could potentially result in negative societal impacts.

\section*{Data Availability Statement}
We confirm that the data supporting the findings of this study are available within published articles of COCO~\cite{lin2014microsoft}, LVIS~\cite{gupta2019lvis}, Visual Genome~\cite{krishna2017visual}, RefCOCO~\cite{yu2016modeling}, RefCOCO+~\cite{yu2016modeling}, and RefCOCOg~\cite{mao2016generation}.

{
\bibliographystyle{unsrt}
\bibliography{reference}
}
\end{document}